\pdfoutput=1
\documentclass{article}

\PassOptionsToPackage{numbers,sort&compress}{natbib}
\usepackage[preprint]{neurips_2026}
\setcitestyle{numbers,square,comma}

\usepackage[utf8]{inputenc}
\usepackage[T1]{fontenc}
\usepackage{textcomp}
\usepackage{microtype}

\usepackage{amsmath}
\usepackage{amsfonts}
\usepackage{nicefrac}
\usepackage{pifont}

\usepackage{graphicx}
\usepackage{booktabs}
\usepackage{longtable}
\usepackage{multirow}
\usepackage[table]{xcolor}
\usepackage{colortbl}
\usepackage{tabularx}
\usepackage{wrapfig}
\usepackage{float}
\usepackage{placeins}
\usepackage{changepage}
\usepackage{rotating}
\usepackage{subcaption}
\usepackage{caption}
\usepackage{tikz}

\usepackage{url}
\usepackage{enumitem}
\usepackage{listings}
\usepackage{tcolorbox}
\usepackage[normalem]{ulem}
\usepackage{ragged2e}
\usepackage{hyperref}

\newcommand{\cmark}{\ding{51}}
\newcommand{\xmark}{\ding{55}}
\newcommand{\pmark}{$\triangle$}
\definecolor{oursbg}{HTML}{EEF4FA}

\title{Faithful, Enriched, and Precise: Benchmarking Natural-Science Illustration Generation by T2I models}

\author{%
  \textbf{Yifan Chang}$^{1,2,3}$ \quad
  \textbf{Jiaxin Ai}$^{1,2,4}$ \quad
  \textbf{Jianwen Sun}$^{1,5}$ \quad
  \textbf{Yuandong Pu}$^{2,6}$ \quad
  \textbf{Siqi Luo}$^{2,6}$ \quad \\
  \textbf{Liangliang Zhao}$^{2,7}$ \quad 
  \textbf{Yuchen Ren}$^{2}$ \quad
  \textbf{Minghao Liu}$^{8}$ \quad
  \textbf{Yunfei Yu}$^{8}$ \quad \\
  \textbf{Yu Qiao}$^{2}$ \quad
  \textbf{Kaipeng Zhang}$^{1,9\dagger}$ \quad
  \textbf{Yihao Liu}$^{2\dagger}$ \quad
  \\[2mm]
    $^1$ Shanghai Innovation Institute \quad
    $^2$ Shanghai AI Laboratory \\
    $^3$ University of Science and Technology of China \quad
    $^4$ Wuhan University \\
    $^5$ Nankai University \quad
    $^6$ Shanghai Jiao Tong University \quad
    $^7$ Fudan University \\
    $^8$ ZODA \quad
    $^9$ Alaya Studio \\
    $^\dagger$ Corresponding authors. \\
    \vspace{-5mm}
}

\begin{document}

\maketitle

\begin{abstract}
  Scientific illustrations are essential tools for communicating research findings, especially in natural science, where they visualize complex concepts and processes. 
As Text-to-Image (T2I) models become increasingly capable, researchers have started to use them for scientific illustration generation. 
However, existing benchmarks often assess outputs at a holistic level, overlooking fine-grained elements, while scientific reasoning ability and output conciseness remain under-quantified. 
We introduce \textbf{FEPBench}, a benchmark built from carefully selected high-quality scientific illustrations across multiple disciplines and layout types. 
With the assistance of multimodal large language models (MLLMs) and human experts, we provide fine-grained atom set annotations and systematically evaluate T2I models along three dimensions: instruction faithfulness, reasoning enrichment, and semantic precision. 
Our evaluation further decomposes model performance across visual, textual, relation, and layout elements. 
Results show that even state-of-the-art (SOTA) closed-source models, such as GPT Image 2 and Nano Banana Pro, still suffer from text-rendering bottlenecks, limited reasoning enrichment, and difficulty balancing generation  richness with precision. 
These findings provide practical guidance for improving and deploying T2I models in scientific illustration generation. Benchmark data, atom set annotations, and evaluation code will be released by us.
\end{abstract}

\vspace{-0.5cm}
\section{Introduction}

Scientific illustrations play a central role in communicating research findings:
they translate complex concepts, mechanisms, and processes into visual
representations that can be inspected, compared, and reasoned about.  Rather than ordinary image generation, scientific illustration generation is a task of structured visual explanation. This role
echoes Marr's view of vision as the construction of useful representations rather
than the mere reproduction of images~\citep{marr1982vision}. Yet for scientific
figures, usefulness is inseparable from communicative clarity: effective graphics
should reveal information rather than obscure it with visual clutter
\citep{tufte1983visual}. These perspectives suggest that scientific figure
generation should not be evaluated only by global visual plausibility. A generated
figure must faithfully realize the intended scientific content, organize it into
a readable visual explanation, and avoid unsupported additions that may distort
scientific communication. Recent text-to-image models, including GPT-Image-2 and Nano Banana Pro, have
made substantial progress in instruction following, text rendering, and visual
reasoning~\citep{openai2026gptimage2,google2025nanobananapro}. This progress has
motivated growing interest in benchmarking scientific and academic figure
generation\cite{zhu2026paperbanana,zhu2026autofigure,chang2025sridbench,lin2026scigenbench},
as well as broader evaluations of knowledge-intensive and reasoning-based image
generation using VQA, graph representations, checklist-based judging, or
multimodal evaluators~\citep{luo2025mmmg,li2026bizgeneval,sun2025t2ireasonbench}.


However, existing protocols remain insufficient for evaluating natural-science
paper figures. Many evaluations rely on holistic multimodal judgments or
high-level scoring prompts, making it difficult to diagnose fine-grained failures
in scientific labels, visual entities, inter-entity relations, and layout
structure. More importantly, they often conflate prompt faithfulness with
reference reconstruction: elements explicitly required by the prompt, details
present only in the reference figure, and unsupported additions in the generated
figure are not clearly distinguished. This distinction is essential in realistic scientific figure generation.
Researchers may provide a concise caption-like prompt describing the main
scientific idea, or use an LLM to reorganize the same content into a structured
visual specification. In either case, content explicitly mentioned in the prompt
should be treated as required; omitting it indicates a failure of instruction
following and scientific understanding. At the same time, a reference figure may
contain additional details that are not explicitly requested. Some of these
details are scientifically meaningful and can enrich the generated figure by
clarifying mechanisms or improving explanatory completeness, while others may be
unnecessary, unsupported, or misleading if generated without grounding in the
prompt or reference annotation. Evaluation should therefore reward faithful
realization of required content, recognize useful optional enrichment, and
penalize unexpected semantic overgeneration.
\vspace{-0.25cm}
\begin{figure}[htbp]  
\centering
\includegraphics[width=1\textwidth]{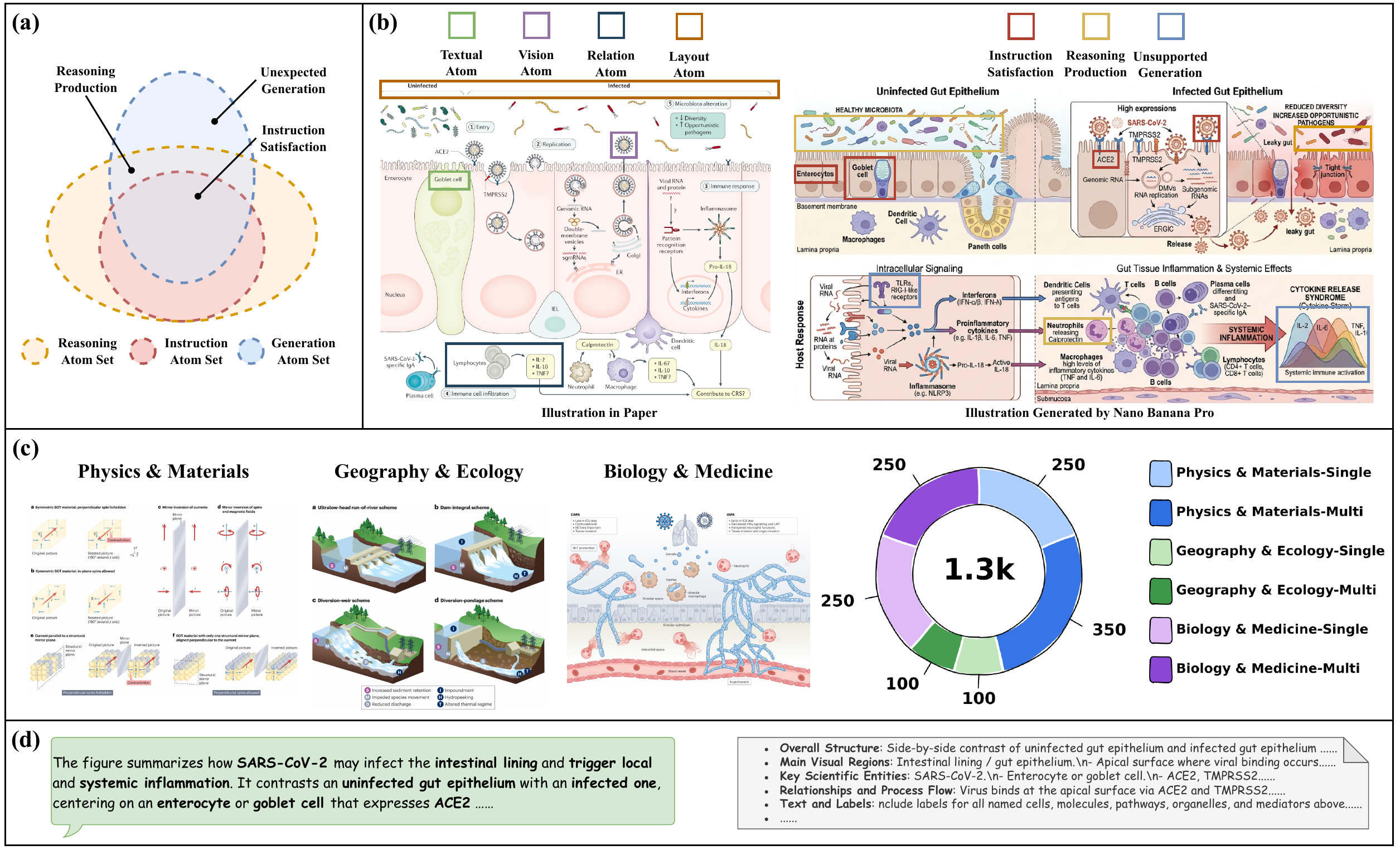} 
\caption{ Overview of our benchmark. (a) Visualization of the atom-set representation. 
(b) Examples of four atom types and three atom-set states. 
(c) Disciplines and data distribution in the benchmark, where single denotes single-panel illustrations and multi denotes multi-panel illustrations. 
(d) Two prompt formats: free-form prompts and structured  prompts.} 
\label{our_bench}   
\end{figure}
\vspace{-0.25cm}

To address this challenge, we propose an atomic-set-based benchmark  \textbf{FEPBench} (\textbf{F}aithful, \textbf{E}nriched, \textbf{P}recise) for
natural-science scientific figure generation. We represent each figure as a set
of semantic atoms, including text atoms, visual atoms, relation atoms, and layout
atoms. Scoring based on fine-grained atom sets is more suitable for scientific illustrations than conventional caption-level or image-level scoring. 
Scientific figures are typically composed of many small elements, and an error in any key element may distort the intended scientific meaning. The gold atom set is partitioned into instruction and reasoning atoms, enabling
us to separate prompt-mandated content from reference-only supplementary details.
Given a generated figure, each atom is assigned to one of three outcome states:
instruction satisfaction, reasoning production, and unexpected
generation. These states induce three complementary evaluation dimensions: Instruction
Faithfulness, which measures fulfillment of required scientific content;
Reasoning Enrichment, which captures recovery of useful optional content; and
Semantic Precision, which penalizes unsupported or excessive generation. Based on
this formulation, our benchmark provides fine-grained, interpretable, and
redundancy-aware evaluation across materials science, earth science, and life
science, covering both single-panel and multi-panel illustrations under
free-form and structured prompt regimes. By comparing free-form and structured prompt regimes, FEPBench further examines whether current T2I models benefit from explicit visual specifications and whether such specifications reduce omission, hallucination, or layout errors.

Overall, our contributions are threefold:
\begin{itemize}    \item We introduce \textbf{FEPBench}, a benchmark of 1,300 high-quality scientific illustrations from natural-science papers, covering three major disciplines and two layout types. For each instance, we provide two types of generation prompts and human-expert-audited atom-set annotations, spanning visual, textual, relational, and layout elements as shown in Fig.\ref{our_bench}.    
\item An evaluation pipeline is proposed which enables fine-grained assessment of generated illustration across different element types along three key dimensions: instruction faithfulness (IF), reasoning enrichment (RE), and semantic precision (SP). The pipeline is robust across different evaluator models and shows strong agreement with human judgments.    
\item We conduct a comprehensive evaluation of SOTA closed-source and open-source T2I models. Our results show that current models still have substantial room for improvement across all three metrics, with text rendering and scientific reasoning emerging as the main bottlenecks.\end{itemize}

\vspace{-0.5cm}

\section{Related Work}
\vspace{-0.3cm}
\paragraph{Text-to-image benchmarks and evaluation.}
Recent T2I evaluation has moved from global image quality and image--text similarity toward more diagnostic tests of faithfulness, compositionality, and reasoning. TIFA~\citep{hu2023tifa} and DSG~\citep{cho2024dsg} decompose prompts into question-answerable or dependency-aware semantic units, while T2I-CompBench~\citep{huang2023t2icompbench}, VQAScore/GenAI-Bench~\citep{lin2024vqascore}, ConceptMix~\citep{wu2024conceptmix}, Gecko~\citep{wiles2025gecko}, and GeckoNum~\citep{kajic2024geckonum} evaluate increasingly fine-grained failures in attribute binding, object relations, compositional control, human-aligned scene understanding, and numerical reasoning. More recent benchmarks further stress reasoning and knowledge-grounded generation: T2I-CoReBench~\citep{li2026t2icorebench} and T2I-ReasonBench~\citep{sun2025t2ireasonbench} test whether models can infer implicit concepts and satisfy reasoning-dependent prompts, while MMMG~\citep{luo2025mmmg} and BizGenEval~\citep{li2026bizgeneval} extend evaluation to multidisciplinary knowledge images and structured commercial visual content. These efforts are especially timely as frontier models such as GPT-Image-2~\citep{openai2026gptimage2} and Nano Banana Pro~\citep{google2025nanobananapro} substantially improve instruction following, text rendering, and visual reasoning. However, most existing protocols still target natural images, general compositional prompts, or broad document-style visual artifacts; they rarely treat natural-science paper figures as a distinct scientific communication medium requiring simultaneous correctness of labels, entities, relations, layout, and semantic precision.

\vspace{-0.25cm}

\begin{table*}[htbp]
\centering
\scriptsize
\setlength{\tabcolsep}{4.2pt}
\renewcommand{\arraystretch}{1.08}
\caption{
Comparison with existing T2I and scientific illustration benchmarks.
\cmark indicates explicit support, \pmark indicates partial or indirect support,
and \xmark indicates that the factor is not explicitly evaluated.
}
\label{tab:benchmark_comparison}
\resizebox{\linewidth}{!}{
\begin{tabular}{lcccccc}
\toprule
Benchmark
& Image Type
& Fine-grained Eval.
& Reasoning Ability
& Conciseness
& Prompt Effect
& Layout Effect
\\
\midrule

T2I-ReasonBench~\citep{sun2025t2ireasonbench} 
& General reasoning images
& \pmark
& \cmark
& \xmark
& \xmark
& \xmark
\\

MMMG-Bench~\citep{luo2025mmmg} 
& Knowledge images
& \cmark
& \cmark
& \xmark
& \xmark
& \xmark
\\

PaperBananaBench~\citep{zhu2026paperbanana}
& Academic method diagrams
& \xmark
& \pmark
& \cmark
& \xmark
& \xmark
\\

FigureBench~\citep{zhu2026autofigure}
& Scientific illustrations
& \xmark
& \pmark
& \xmark
& \xmark
& \pmark
\\

SridBench~\citep{chang2025sridbench}
& Scientific illustrations
& \xmark
& \xmark
& \pmark
& \xmark
& \xmark
\\

\rowcolor{oursbg}
\textbf{FEPBench (ours)}
& \textbf{Scientific illustrations}
& \cmark
& \cmark
& \cmark
& \cmark
& \cmark
\\

\bottomrule
\end{tabular}
}
\end{table*}
\vspace{-0.2cm}

\paragraph{Scientific and academic illustration generation.} Scientific figure generation has recently emerged as a specialized form of visual synthesis, where the goal is not only to produce plausible images but to communicate mechanisms, processes, and structures with scientific clarity. In the broader structured-visual domain, DiagramAgent~\citep{wei2025diagramagent} studies text-to-diagram generation and editing for flowcharts, model architectures, and other editable diagrams. More directly, PaperBanana~\citep{zhu2026paperbanana} automates academic illustration through an agentic pipeline for reference retrieval, planning, rendering, and self-critique; AutoFigure~\citep{zhu2026autofigure} introduces FigureBench and an agentic framework for generating and refining publication-ready scientific illustrations from scientific text; SridBench~\citep{chang2025sridbench} benchmarks scientific  illustration drawing across disciplines; and SciGenBench~\citep{lin2026scigenbench} evaluates scientific image synthesis by downstream reasoning utility and logical validity. These studies establish scientific illustration generation as an important emerging task, but their evaluation is still limited for natural-science illustrations: they either emphasize holistic multimodal judgment or general scientific utility, without explicitly separating prompt-required content from reference-only enrichment or penalizing unsupported semantic additions. In contrast, our work formulates evaluation as atomic-set matching over text, visual, relation, and layout atoms, enabling fine-grained measurement of prompt faithfulness, useful reasoning enrichment, and redundancy-aware semantic precision. As shown in Tab.\ref{tab:benchmark_comparison}, FEPBench advances prior work in multiple aspects and introduces several  innovations.

\vspace{-0.25cm}

\section{Benchmark Overview}
\label{sec:benchmark}
\vspace{-0.3cm}
We collected 1,300 illustrations through expert screening under strict inclusion criteria. 
Each image must be clear, contain both visual and textual elements, and exclude experimental results, data visualizations, and imaging outputs. 
To ensure scientific validity and semantic richness, the figures were sourced from authoritative journals under Nature Portfolio and used with appropriate authorization for non-commercial research evaluation. We intentionally avoid disclosing detailed source lists in the manuscript and provide only the necessary high-level provenance information.
The dataset covers three broad domains: physics and materials, geography and ecology, and biology and medicine. 
For prompt construction, we invited human experts with PhD degrees in relevant fields to write free-form textual prompts based on the original paper context and the corresponding illustration, following how users would naturally instruct a T2I model. 
We then used an LLM (GPT-5.4) to convert these prompts into structured prompts. 
During this conversion, the LLM was explicitly instructed to preserve the original meaning and not to add, remove, or alter any visual, textual, relational, or layout elements. We also annotate the aspect ratio of each reference illustration, including landscape, portrait, and square formats, to facilitate selecting a generation resolution closer to the reference figure.

As shown in Fig.\ref{our_pipeline}, we use an MLLM to extract the corresponding atom set from each selected illustration. 
An OCR model is used to extract in-figure text, which serves as the reference for textual element detection. 
After obtaining the atom sets, human experts inspect and correct each atom set against the corresponding illustration. 
The detailed prompts are provided in the Appendix Sec.\ref{prompt}.

After obtaining the outputs from T2I models, we use an MLLM to evaluate them against the atom-set file of the reference illustration. 
The evaluator assesses the realization of both instruction atoms and reasoning atoms, and also detects unexpected atoms that are not supported by the reference or prompt. 
Based on these judgments, we compute the final evaluation metrics according to our definitions.

\vspace{-0.25cm}
\section{Atomic Set-Based Evaluation}

\vspace{-0.45cm}

\begin{figure}[htbp]  
\centering
\includegraphics[width=1\textwidth]{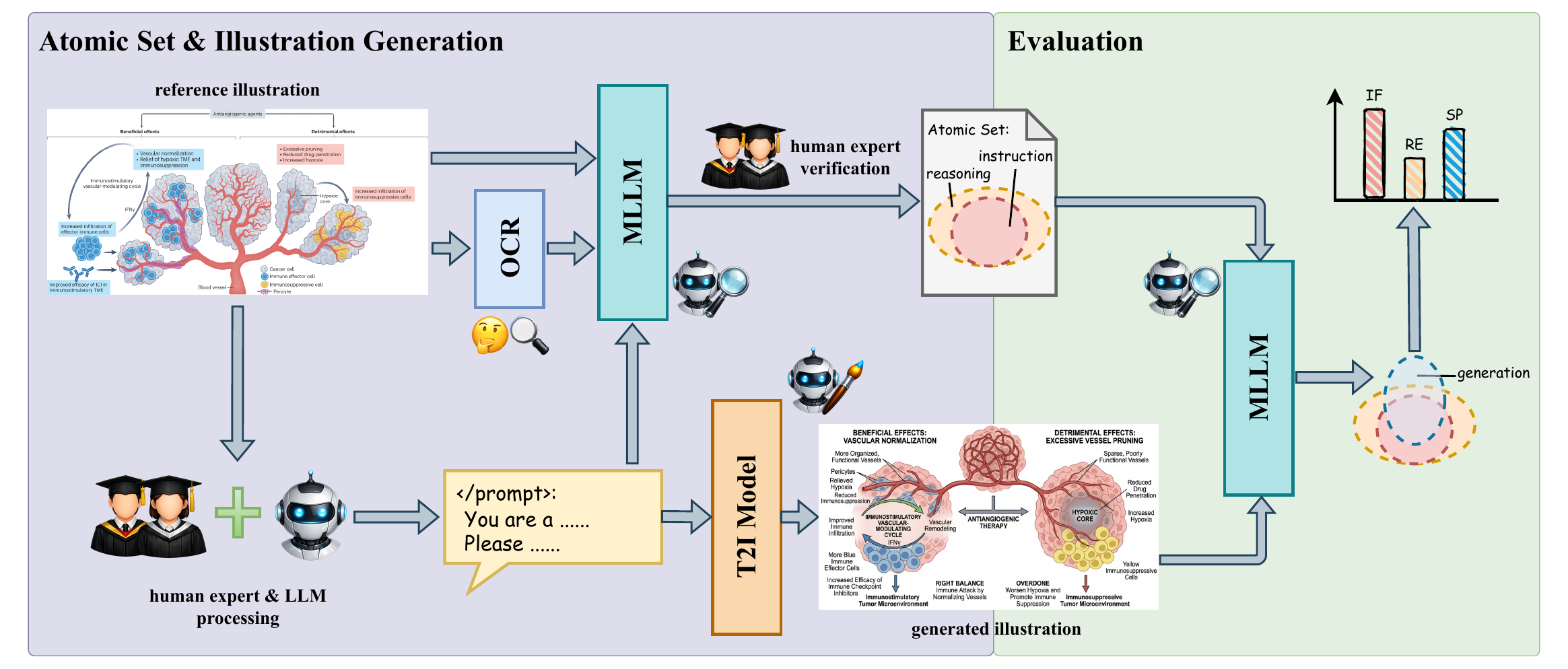} 
\caption{ Pipeline of our benchmark } 
\label{our_pipeline}   
\end{figure}
\vspace{-0.5cm}
\paragraph{Definition of Atom Set.}
We evaluate generated scientific illustrations via a state-based accounting framework over atoms. Let $\mathcal{A}$ denote the gold  atom set of a target illustration, partitioned into instruction atoms $\mathcal{A}^{ins}$ and reasoning atoms $\mathcal{A}^{rea}$, and let $\hat{\mathcal{A}}$ denote the realized atom set of a generated illustration. $a\in\mathcal{A}^{ins}$ means an atom that can find evidence in instructions (prompts), while $a\in\mathcal{A}^{rea}$ is an atom that can only find in the reference illustration.  Given $\hat{\mathcal{A}}$, each atom falls into one of three outcome states: \emph{instruction satisfaction} $\left(\mathcal{A}^{ins}\cap\hat{\mathcal{A}}\right)$, \emph{reasoning production} $\left(\mathcal{A}^{rea}\cap\hat{\mathcal{A}}\right)$, and \emph{unexpected generation} $\left(\hat{\mathcal{A}}\setminus(\mathcal{A}^{ins}\cup\mathcal{A}^{rea})\right)$. These three states induce three complementary evaluation dimensions: \emph{Instruction Faithfulness} (IF), \emph{Reasoning Enrichment} (RE), and \emph{Semantic Precision} (SP).

As shown in Fig.\ref{our_pipeline}, our benchmark is constructed through an OCR and MLLM-assisted atomic annotation and verification pipeline.
Given a reference  illustration $I$ and its generation prompt $p$,  OCR is applied to extract candidate textual elements from $I$ and then we use  MLLM to jointly analyze the reference illustration and prompt.
This step produces a gold semantic atom set $\mathcal{A}=\mathcal{A}^{ins}\cup\mathcal{A}^{rea}$.

For generation, each target illustration is paired with two prompt regimes: a free-form expert prompt $p^{free}$ and an LLM-organized structured prompt $p^{struct}$.
Each T2I model $G$ generates two outputs,
\begin{equation}
\hat{I}^{free}=G(p^{free}),\qquad
\hat{I}^{struct}=G(p^{struct}).
\end{equation}
Given a generated figure $\hat{I}$, we again apply  an MLLM verifier to produces a realized atom set $\hat{\mathcal{A}}$ and check it against the gold atom set.

\vspace{-0.2cm}

\paragraph{Atom-level matching.}
In our implementation, semantic atoms are instantiated as textual, visual, relation, and layout atoms, yielding
\begin{equation}
\mathcal{A}=\mathcal{A}_t \cup \mathcal{A}_v \cup \mathcal{A}_r \cup \mathcal{A}_l,
\end{equation}
where $\mathcal{A}_t$, $\mathcal{A}_v$, $\mathcal{A}_r$, and $\mathcal{A}_l$ denote textual, visual, relation, and layout atoms, respectively. For each text atom $a \in \mathcal{A}_t$, we define a matching function 
\begin{equation}
m_t(a)
=
\mathrm{Exact}(a)\cdot\Bigl(
\beta \cdot\,\mathrm{Read}(a)
+
\beta\cdot\,\mathrm{Attach}(a) \Bigl)
\end{equation}

$\mathrm{Exact}(a)\in\{0,1\}$ is indicated as textual exact equivalence, $\mathrm{Read}(a)\in\{0,1\}$ indicates readability, and $\mathrm{Attach}(a)\in\{0,1\}$ indicates correct attachment to intended object or region. $\beta=0.5$  is normalization factors.  Similarly, for each visual atom $a \in \mathcal{A}_v$, we define
\begin{equation}
m_v(a)
=
\mathrm{Pres}(a)\cdot
\Bigl(
\beta\cdot\,\mathrm{Cnt}(a)
+
\beta\cdot\,\mathrm{Loc}(a) \Bigl)
\end{equation}
where $\mathrm{Pres}(a)\in\{0,1\}$ corresponds to absent/present status, $\mathrm{Cnt}(a)$ and $\mathrm{Loc}(a)$ denote count, and coarse-location correctness. For each relation atom $a \in \mathcal{A}_r$, we define
\begin{equation}
m_r(a)
=
\mathrm{Stat}(a)\cdot
\Bigl(
\beta\cdot\,\mathrm{Type}(a)
+
\beta\cdot\,\mathrm{Dir}(a)
\Bigr),
\end{equation}

Here, $\mathrm{Stat}(a)\in\{0,1\}$ represents violated/satisfied status, $\mathrm{Type}(a)\in\{0,1\}$ indicates relation-type correctness, $\mathrm{Dir}(a)\in\{0,1\}$ indicates directional correctness.

For each layout atom $a \in \mathcal{A}_l$, we use a binary local match $m_l(a)\in\{0,1\}$, where $m_l(a)=1$ if the corresponding layout constraint is satisfied and $m_l(a)=0$ otherwise. In our implementation, the layout atoms correspond to panel-structure match, panel-count match, and global-layout match.
\vspace{-0.2cm}
\paragraph{Instruction Faithfulness and Reasoning Enrichment.}
For each semantic modality $x \in \{t,v,r\}$ with  matching function $m_x(\cdot)$, we define
\begin{equation}
IF_x
=
\frac{1}{|\mathcal{A}_x^{ins}|}
\sum_{a\in\mathcal{A}_x^{ins}} m_x(a),\  IF_l
=
\frac{1}{|\mathcal{A}_l^{ins}|}
\sum_{a\in\mathcal{A}_l^{ins}} m_l(a).\ 
RE_x
=
\frac{1}{|\mathcal{A}_x^{rea}|}
\sum_{a\in\mathcal{A}_x^{rea}} m_x(a)
\end{equation}

\begin{equation}
\mathrm{IF}
=
\frac{1}{4}
\left(
IF_v
+
IF_t
+
IF_r
+
IF_l
\right),
\qquad
\mathrm{RE}
=
\frac{1}{3}
\left(
RE_v
+
RE_t
+
RE_r
\right).
\end{equation}
Thus, $\mathrm{IF}$ measures how well the generated figure satisfies prompt-required scientific content, whereas $\mathrm{RE}$ measures how much scientifically useful reference-only content is additionally recovered. It is worth noting that, to make the metrics computable, we require each sample to contain a non-zero number of atoms in every category during data collection. The average number of atoms in each category for each illustration type is reported in the Appendix Tab.\ref{tab:atom_statistics_by_subset}.
\vspace{-0.2cm}
\paragraph{Semantic Precision.}
We define unsupported scientific content as semantic atoms realized in the generated figure that align to neither required nor optional atoms. Let $\hat{\mathcal{A}}_{x}^{ex}$ and $\hat{\mathcal{A}}_{x}^{unex}$ denote the sets of supported and unsupported realized atoms, respectively. We then define
\begin{equation}
SP_x = 1-\frac{
\left|\hat{\mathcal{A}}_{x}^{unex}\right|
}{
\left|\hat{\mathcal{A}}_{x}^{ex}\right|
+
\left|\hat{\mathcal{A}}_{x}^{unex}\right|
},\qquad \mathrm{SP}= \frac{1}{3}
\left(
SP_v
+
SP_t
+
SP_r
\right).
\end{equation}
A higher $\mathrm{SP}$ indicates that the realized semantic content remains more strongly grounded in prompt-supported or reference-supported scientific structure, rather than drifting into unsupported additions.


\definecolor{closedbg}{HTML}{EEF4FA}
\definecolor{openbg}{HTML}{FBF4E6}
\definecolor{promptbg}{HTML}{E8E8E8}

\newcommand{\best}[1]{\textbf{#1}}
\newcommand{\second}[1]{\underline{#1}}

\begin{table*}[htbp]
\centering
\scriptsize
\setlength{\tabcolsep}{2.2pt}
\renewcommand{\arraystretch}{1.05}
\caption{
Scores of different models on each atom type and the overall score.
Best and second-best results are shown in \textbf{bold} and \underline{underline}, respectively.
Closed-source and open-source models are shaded with different background colors.
}
\label{tab:finegrained_model_metrics}
\resizebox{\linewidth}{!}{
\begin{tabular}{lccccccccccccc}
\toprule
&
\multicolumn{4}{c}{Instruction Faithfulness}
&
\multicolumn{3}{c}{Reasoning Enrichment}
&
\multicolumn{3}{c}{Semantic Precision}
&
\multicolumn{3}{c}{Mean}
\\
\cmidrule(lr){2-5}
\cmidrule(lr){6-8}
\cmidrule(lr){9-11}
\cmidrule(lr){12-14}
Model
& $IF_{v}\uparrow$
& $IF_{t}\uparrow$
& $IF_{r}\uparrow$
& $IF_{\ell}\uparrow$
& $RE_{v}\uparrow$
& $RE_{t}\uparrow$
& $RE_{r}\uparrow$
& $SP_{v}\uparrow$
& $SP_{t}\uparrow$
& $SP_{r}\uparrow$
& $IF\uparrow$
& $RE\uparrow$
& $SP\uparrow$
\\
\midrule

\rowcolor{closedbg}
GPT Image 2
& \best{0.809}
& \best{0.389}
& \best{0.573}
& \best{0.886}
& \best{0.417}
& \best{0.141}
& \best{0.443}
& 0.774
& 0.479
& \second{0.856}
& \best{0.664}
& \best{0.334}
& 0.703
\\

\rowcolor{closedbg}
Nano Banana Pro
& 0.772
& 0.333
& \second{0.541}
& \second{0.870}
& 0.329
& \second{0.119}
& 0.395
& 0.812
& \second{0.507}
& \best{0.879}
& \second{0.629}
& \second{0.281}
& \second{0.733}
\\

\rowcolor{closedbg}
GPT Image 1.5
& \second{0.793}
& 0.361
& 0.529
& 0.710
& \second{0.344}
& 0.099
& \second{0.438}
& \second{0.859}
& 0.438
& 0.784
& 0.598
& 0.294
& 0.693
\\

\rowcolor{closedbg}
Seedream 5.0
& 0.748
& \second{0.362}
& 0.505
& 0.685
& 0.266
& 0.100
& 0.398
& \best{0.869}
& 0.485
& 0.793
& 0.575
& 0.255
& 0.716
\\

\rowcolor{closedbg}
Qwen-Image-2.0 Pro
& 0.751
& 0.234
& 0.432
& 0.641
& 0.277
& 0.054
& 0.371
& 0.852
& \best{0.565}
& 0.824
& 0.515
& 0.234
& \best{0.747}
\\

\rowcolor{openbg}
FLUX.2 [dev]
& 0.724
& 0.289
& 0.424
& 0.686
& 0.306
& 0.058
& 0.368
& 0.832
& 0.320
& 0.812
& 0.531
& 0.244
& 0.655
\\

\rowcolor{openbg}
HunyuanImage-3.0
& 0.632
& 0.087
& 0.265
& 0.655
& 0.293
& 0.029
& 0.279
& 0.746
& 0.295
& 0.798
& 0.410
& 0.200
& 0.613
\\

\rowcolor{openbg}
Z-Image-Turbo
& 0.560
& 0.179
& 0.259
& 0.562
& 0.232
& 0.030
& 0.234
& 0.791
& 0.298
& 0.730
& 0.390
& 0.165
& 0.606
\\

\rowcolor{openbg}
Ovis-Image
& 0.478
& 0.101
& 0.173
& 0.584
& 0.174
& 0.023
& 0.179
& 0.825
& 0.230
& 0.800
& 0.334
& 0.125
& 0.618
\\

\bottomrule
\end{tabular}
}
\end{table*}
\section{Experiment}
\vspace{-0.2cm}
\paragraph{Models and generation protocol.}
We benchmark nine text-to-image models spanning proprietary frontier systems and recent open-source generators: GPT Image 2~\citep{openai2026gptimage2}, Nano Banana Pro~\citep{google2026nanobananapro}, GPT Image 1.5~\citep{openai2025gptimage15}, Qwen-Image-2.0 Pro~\citep{qwen2026qwenimage20,alibabacloud2026qwenmodels}, Seedream 5.0~\citep{bytedance2026seedream50}, FLUX.2 [dev]~\citep{blackforestlabs2025flux2}, Z-Image-Turbo~\citep{zimage2025technical}, Ovis-Image~\citep{wang2025ovisimage}, and HunyuanImage-3.0~\citep{cao2025hunyuanimage3}. 
All models are evaluated with identical prompts under the same benchmark split, covering both natural and structured prompt settings. 
We do not use model-specific prompt engineering, manual image selection, post-generation editing, or any human intervention. 
For open-source models, we use the official checkpoints and recommended inference settings; for API-based models, we use their default generation configurations. 
When supported, we fix the random seed to 42; otherwise, we rely on the model's default stochastic generation process. 
Output resolution and aspect ratio are matched as closely as each model allows to the reference figure. Details can be
 seen in Tab.\ref{tab:generation_settings}
\vspace{-0.2cm}
\paragraph{Evaluation pipeline.}
We build gold atom sets with an OCR--MLLM pipeline. 
We first use MinerU-Diffusion-V1-0320-2.5B~\citep{dong2026minerudiffusion}, locally deployed on a single NVIDIA RTX 4090 GPU, to extract textual labels and layout cues from each reference illustration. 
GPT-5.4~\citep{openai2026gpt54} is then used as the multimodal annotator to parse the reference figure, OCR results, and prompt into visual entities, text elements, and semantic relations. 
We use both GPT-5.4 and Qwen3.5-397B-A17B~\citep{qwen2026qwen35} as MLLM evaluators. 
Qwen3.5-397B-A17B is locally deployed, whereas all GPT-5.4 calls are made with temperature fixed to 0. Additional human-expert annotation details and protocols will be released with the benchmark materials.

\vspace{-0.25cm}
\begin{figure}[htbp]  
\centering
\includegraphics[width=1\textwidth]{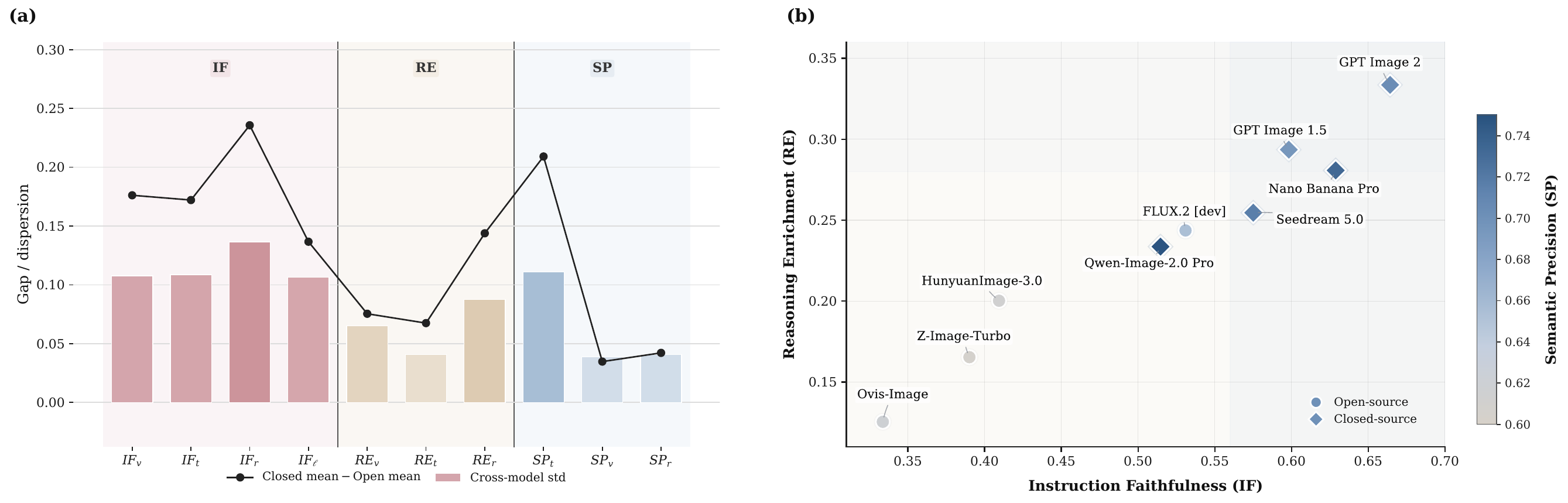} 
\caption{
Overall model comparison from fine-grained metric gaps to the capability frontier.
$\textbf{(a)}$ Cross-model dispersion and the closed--open source performance gap across fine-grained IF, RE, and SP submetrics. Bars denote the standard deviation across models, while the line shows the mean difference between closed-source and open-source models.
$\textbf{(b)}$  Model-level capability frontier in the IF--RE space, with color indicating semantic precision (SP). 
}
\label{frontier}   
\end{figure}
\vspace{-0.35cm}
\paragraph{Main Results} Tab.~\ref{tab:finegrained_model_metrics} highlights  GPT Image 2 substantially outperforms other models in instruction faithfulness and reasoning enrichment across all element types. 
Nano Banana Pro shows a more balanced profile, combining strong faithfulness, enrichment, and precision. 
Qwen-Image-2.0 Pro achieves the best semantic precision. 
Overall, closed-source models consistently outperform open-source models. 
Notably, however, RE remains clearly weaker than the other two dimensions for every model, indicating limited scientific reasoning capability. 
The fine-grained results further show that text-related metrics are substantially lower than other metrics, and that all models perform poorly on the instruction faithfulness of relational atoms. 
These findings suggest that, although SOTA T2I models can render visual elements reasonably well, they still struggle to generate in-figure text and to reconstruct the scientific relations among elements. Additional statistical analyses of the results are provided in Appendix Sec.\ref{stat}.

\definecolor{closedbg}{HTML}{EEF4FA}
\definecolor{openbg}{HTML}{FBF4E6}
\definecolor{promptbg}{HTML}{E8E8E8}

\begin{table*}[htbp]
\centering
\scriptsize
\setlength{\tabcolsep}{2.3pt}
\renewcommand{\arraystretch}{1.06}
\caption{
Fine-grained comparison under free-form and  structured
prompts across three scientific domains.
Overall score is weighted by the number of samples. 
Within each prompt block, the best and second-best results are marked in \textbf{bold} and \underline{underline}, respectively.
Closed-source and open-source models are shaded with different background colors.
}
\label{tab:prompt_subject_finegrained}
\resizebox{\linewidth}{!}{
\begin{tabular}{lcccccccccccc}
\toprule
&
\multicolumn{3}{c}{Physics \& Materials}
&
\multicolumn{3}{c}{Geography \& Ecology}
&
\multicolumn{3}{c}{Biology \& Medicine}
&
\multicolumn{3}{c}{Overall}
\\
\cmidrule(lr){2-4}
\cmidrule(lr){5-7}
\cmidrule(lr){8-10}
\cmidrule(lr){11-13}
Model
& $IF\uparrow$ & $RE\uparrow$ & $SP\uparrow$
& $IF\uparrow$ & $RE\uparrow$ & $SP\uparrow$
& $IF\uparrow$ & $RE\uparrow$ & $SP\uparrow$
& $IF\uparrow$ & $RE\uparrow$ & $SP\uparrow$
\\
\midrule

\rowcolor{promptbg}\multicolumn{13}{l}{\textbf{Free-form prompt}} \\

\rowcolor{closedbg}
GPT Image 2
& \best{0.656} & \best{0.338} & \second{0.738}
& \best{0.643} & \best{0.333} & \second{0.736}
& \best{0.679} & \best{0.328} & 0.682
& \best{0.663} & \best{0.334} & \second{0.716}
\\

\rowcolor{closedbg}
Nano Banana Pro
& \second{0.636} & \second{0.308} & 0.719
& \second{0.608} & \second{0.306} & 0.710
& \second{0.651} & 0.272 & 0.708
& \second{0.638} & \second{0.294} & 0.713
\\

\rowcolor{closedbg}
GPT Image 1.5
& 0.595 & 0.299 & 0.686
& 0.584 & 0.302 & 0.649
& 0.600 & \second{0.276} & 0.705
& 0.595 & 0.290 & 0.688
\\

\rowcolor{closedbg}
Seedream 5.0
& 0.575 & 0.271 & 0.708
& 0.560 & 0.250 & 0.685
& 0.590 & 0.247 & \second{0.725}
& 0.578 & 0.259 & 0.711
\\

\rowcolor{closedbg}
Qwen-Image-2.0 Pro
& 0.491 & 0.228 & \best{0.750}
& 0.489 & 0.263 & \best{0.752}
& 0.527 & 0.237 & \best{0.733}
& 0.505 & 0.237 & \best{0.744}
\\

\rowcolor{openbg}
FLUX.2 [dev]
& 0.509 & 0.231 & 0.675
& 0.489 & 0.257 & 0.648
& 0.532 & 0.234 & 0.667
& 0.515 & 0.236 & 0.668
\\

\rowcolor{openbg}
HunyuanImage-3.0
& 0.356 & 0.169 & 0.619
& 0.335 & 0.140 & 0.682
& 0.370 & 0.181 & 0.557
& 0.358 & 0.169 & 0.605
\\

\rowcolor{openbg}
Z-Image-Turbo
& 0.366 & 0.148 & 0.625
& 0.376 & 0.165 & 0.590
& 0.367 & 0.143 & 0.603
& 0.368 & 0.149 & 0.611
\\

\rowcolor{openbg}
Ovis-Image
& 0.344 & 0.129 & 0.651
& 0.376 & 0.168 & 0.636
& 0.322 & 0.112 & 0.623
& 0.340 & 0.129 & 0.638
\\

\addlinespace[1pt]

\rowcolor{promptbg}\multicolumn{13}{l}{\textbf{Structured prompt}} \\

\rowcolor{closedbg}
GPT Image 2
& \best{0.665} & \best{0.342} & 0.705
& \best{0.651} & \best{0.343} & 0.685
& \best{0.673} & \best{0.320} & 0.675
& \best{0.666} & \best{0.334} & 0.690
\\

\rowcolor{closedbg}
Nano Banana Pro
& \second{0.622} & 0.277 & \second{0.754}
& \second{0.593} & 0.278 & \best{0.753}
& \second{0.628} & 0.251 & \second{0.748}
& \second{0.620} & 0.267 & \best{0.752}
\\

\rowcolor{closedbg}
GPT Image 1.5
& 0.596 & \second{0.304} & 0.697
& 0.587 & \second{0.314} & 0.652
& 0.613 & \second{0.281} & 0.720
& 0.602 & \second{0.297} & 0.699
\\

\rowcolor{closedbg}
Seedream 5.0
& 0.572 & 0.255 & 0.720
& 0.555 & 0.268 & 0.698
& 0.579 & 0.238 & 0.730
& 0.572 & 0.251 & 0.720
\\

\rowcolor{closedbg}
Qwen-Image-2.0 Pro
& 0.513 & 0.224 & \best{0.754}
& 0.504 & 0.254 & \second{0.738}
& 0.548 & 0.229 & \best{0.751}
& 0.525 & 0.231 & \second{0.751}
\\

\rowcolor{openbg}
FLUX.2 [dev]
& 0.535 & 0.250 & 0.641
& 0.531 & 0.278 & 0.618
& 0.568 & 0.242 & 0.652
& 0.547 & 0.251 & 0.642
\\

\rowcolor{openbg}
HunyuanImage-3.0
& 0.476 & 0.239 & 0.625
& 0.374 & 0.188 & 0.691
& 0.478 & 0.238 & 0.588
& 0.461 & 0.231 & 0.621
\\

\rowcolor{openbg}
Z-Image-Turbo
& 0.400 & 0.179 & 0.608
& 0.419 & 0.221 & 0.581
& 0.423 & 0.170 & 0.600
& 0.412 & 0.182 & 0.601
\\

\rowcolor{openbg}
Ovis-Image
& 0.337 & 0.124 & 0.606
& 0.353 & 0.150 & 0.625
& 0.304 & 0.108 & 0.579
& 0.327 & 0.122 & 0.599
\\

\bottomrule
\end{tabular}
}
\end{table*}

As shown in Fig.~\ref{frontier} (a), fine-grained evaluation reveals that the main gaps between models lie in overall instruction faithfulness, especially faithful realization of relational atoms, and the semantic precision of textual elements. 
These dimensions also account for the overall advantage of closed-source models over open-source ones. Fig.~\ref{frontier} (b) shows that in general, the strongest closed-source models, including GPT Image 2, Nano Banana Pro, and GPT Image 1.5, show clear advantages over other models in scientific illustration generation. Among open-source models, FLUX.2 [dev] demonstrates the strongest performance, approaching the capability of some closed-source models.

\vspace{-0.3cm}

\paragraph{Factor-wise Analysis: Subject, Layout, and Prompting} Table~\ref{tab:prompt_subject_finegrained} shows that model rankings are generally stable across scientific subjects. The relative ranking of models remains largely stable: GPT Image 2 consistently achieves the highest $IF$ and $RE$ in all three subjects, while Qwen-Image-2.0 Pro and Nano Banana Pro obtain stronger $SP$, suggesting that semantic coverage and precision are affected by different model capabilities. Domain difficulty is not uniform across metrics. Biology \& Medicine often yields slightly higher $IF$, especially for stronger closed-source models, whereas Geography \& Ecology shows more variability in $SP$, indicating that visually dense ecological and geoscience processes may introduce more unsupported or imprecise elements. The effect of prompt rewriting is also model-dependent. Structured prompts improve $IF$ for several models, especially open-source systems such as FLUX.2 [dev] and Hunyuan-Image-3.0, but can reduce $SP$ by encouraging additional, unsupported details. Thus, prompt rewriting is not a universal improvement strategy; instead, it shifts the coverage--precision trade-off differently across model families and scientific domains.

\begin{figure}[htbp]  
\centering
\includegraphics[width=1\textwidth]{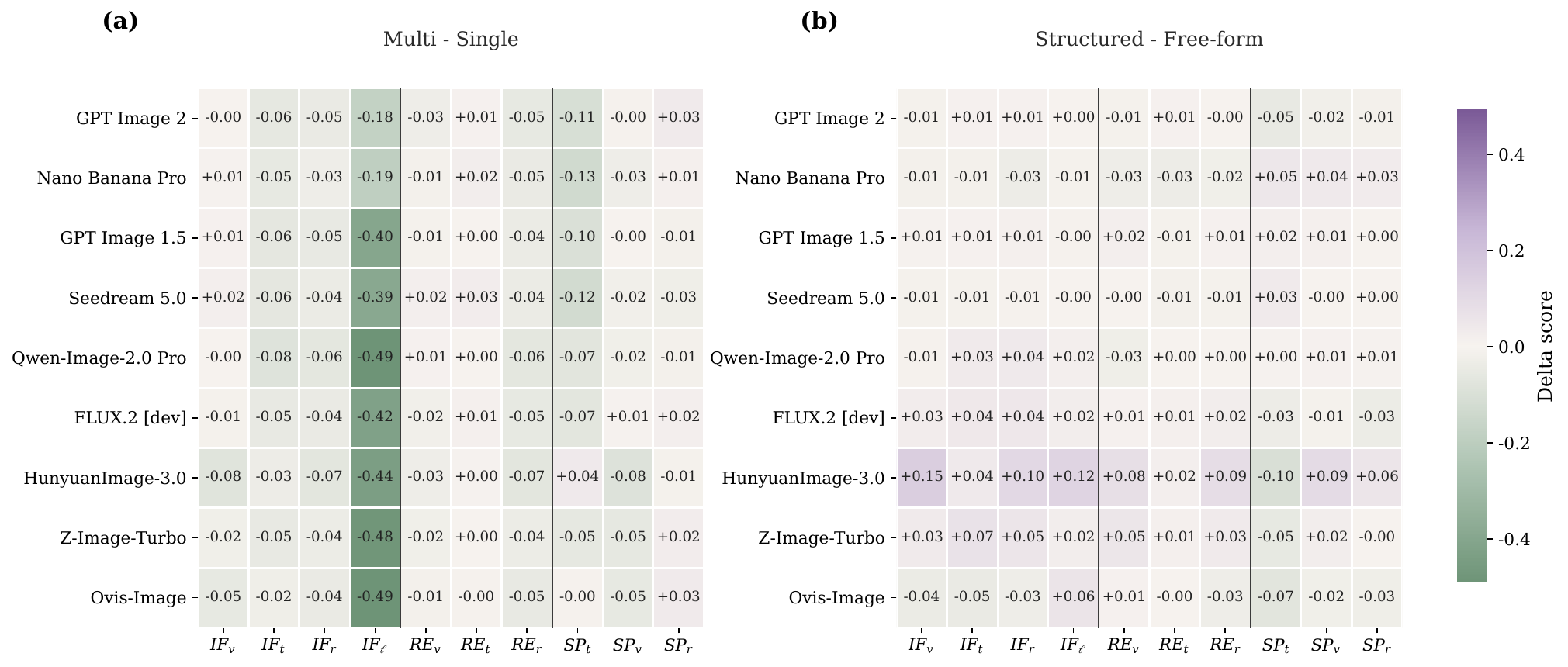} 
\caption{
Effects of illustration layout and prompt format on fine-grained model performance.
$\textbf{(a)}$  Score differences between multi-panel and single-panel illustrations.
$\textbf{(b)}$  Score differences between structured and free-form prompts.
Positive values indicate that the former setting improves performance over the latter.
Columns cover fine-grained IF, RE, and SP submetrics, with darker colors indicating larger condition-induced changes.
}

\label{delta}   
\end{figure}

Fig.\ref{delta} (a) shows, for each model, the mean score difference on multi-panel illustrations relative to single-panel illustrations for each fine-grained metric. 
We observe that, for nearly all models, most scores decrease on multi-panel data. 
The largest drop occurs in layout accuracy, which is the most direct and expected consequence of increased layout complexity. 
Compared with the other dimensions, instruction faithfulness (IF) is more negatively affected, and the IF degradation of open-source models is clearly larger than that of closed-source models. 
Fig.\ref{delta} (b) shows that converting inputs into structured prompts has only a minor effect on closed-source models. 
In contrast, for several open-source models, structured prompts improve overall performance, especially IF. 
These findings suggest that closed-source models are less affected by increased task difficulty and are also more robust to prompt format.

\vspace{-0cm}

\begin{figure}[htbp]  
\centering
\includegraphics[width=1\textwidth]{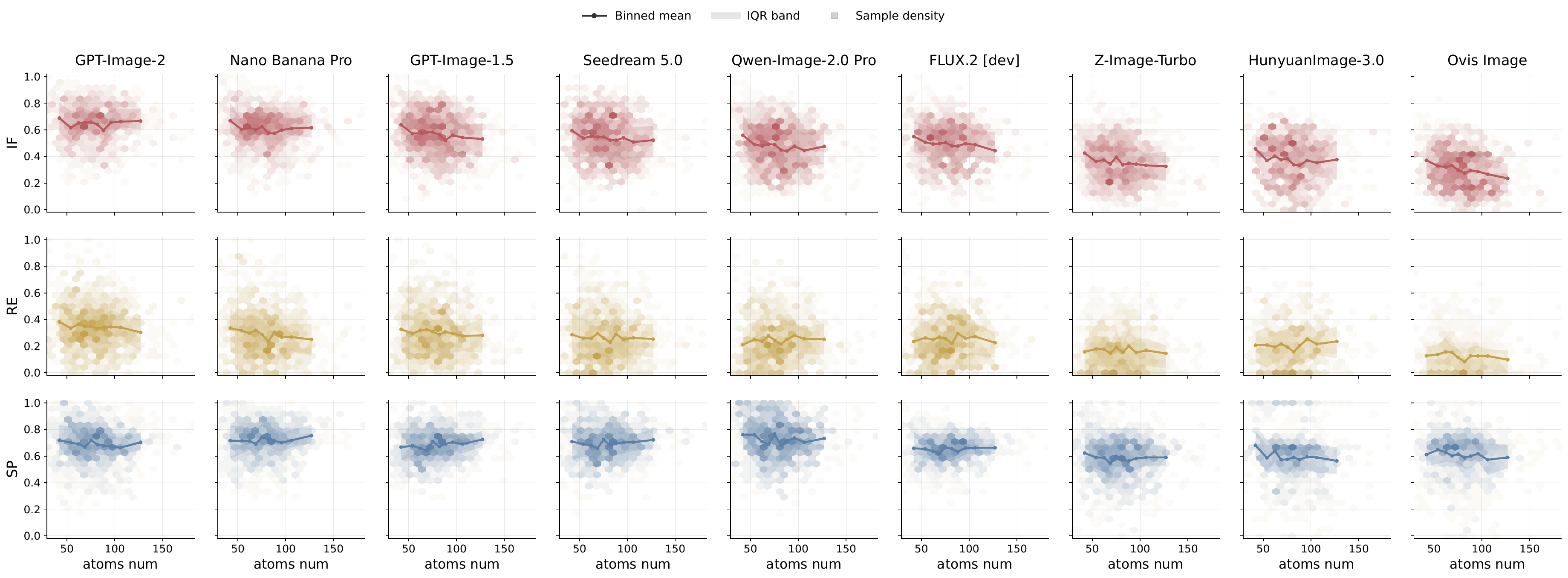} 
\caption{
Model robustness to increasing semantic complexity.
For each model, we plot IF, RE, and SP against the total number of semantic atoms.
Density shows the empirical sample distribution; curves and shaded regions indicate quantile-binned means and IQRs, respectively, where IQR denotes the 25th--75th percentile range.
}

\label{atom}   
\end{figure}

\paragraph{Effect of atom set complexity.} We further analyze how model performance changes as scientific illustrations become compositionally more complex.
We measure complexity by the total number of semantic atoms, covering instruction and reasoning atoms.
As shown in Fig~\ref{atom}, among the three dimensions, IF is most affected by atom-set complexity. 
As the number of atoms increases, the IF scores of all models decline to varying degrees. 
However, GPT Image 2 and Nano Banana Pro exhibit the smallest performance degradation. 
Their sample-level score distributions are also more concentrated than those of other models. 
This indicates that these two models are more robust to increasing semantic complexity.

\vspace{-0.4cm}

\vspace{-0.1cm}

\begin{figure}[htbp]  
\centering
\includegraphics[width=1\textwidth]{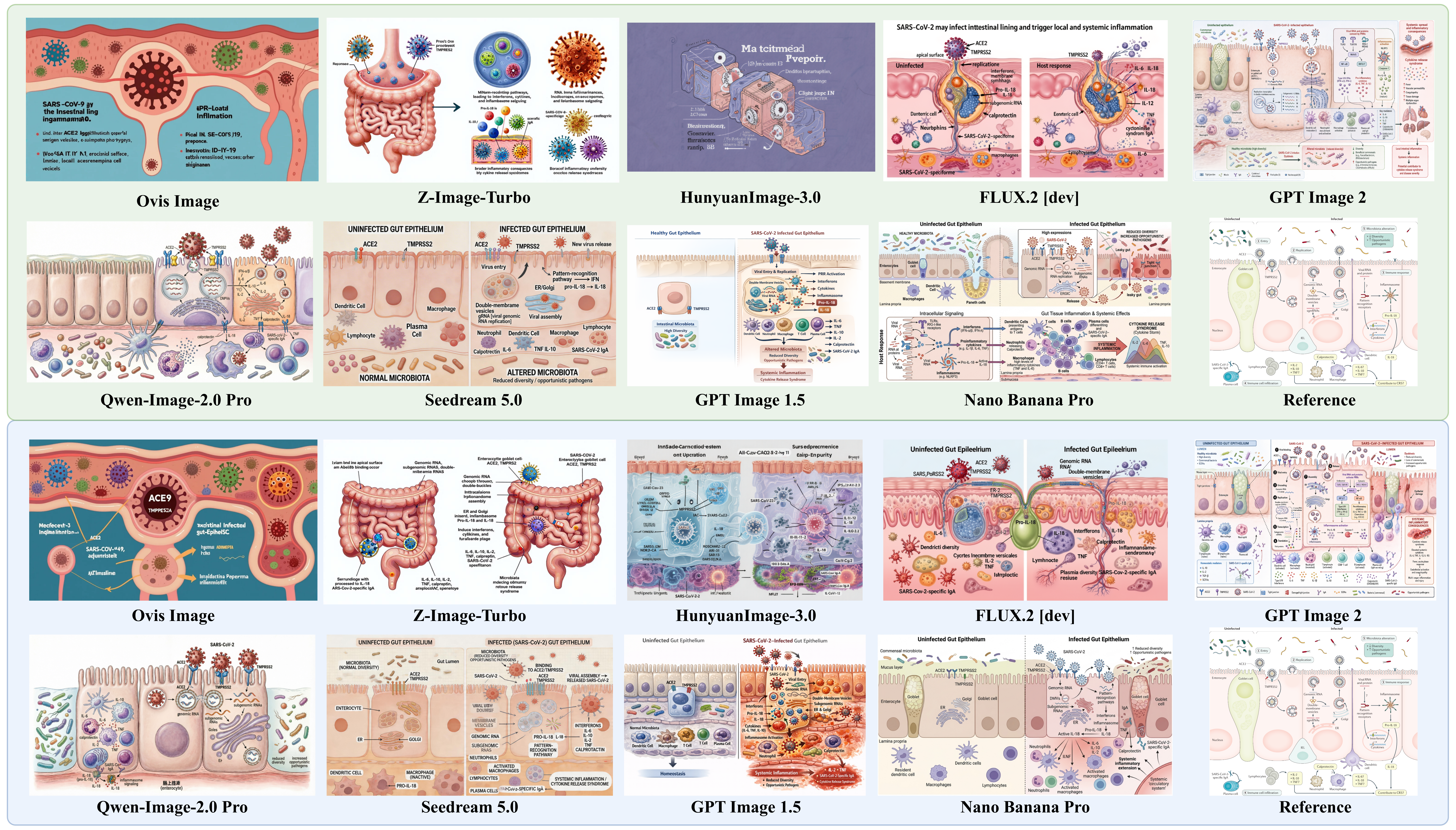} 
\caption{ Comparison of generated results from different models under different prompt formats for the same reference figure. 
The top row uses free-form textual prompts, while the bottom row uses structured prompts.} 
\label{result}   
\end{figure}

Fig~\ref{result} provides a qualitative comparison of model outputs under two prompt formats for the same reference illustration. 
Overall, structured prompts improve the organization and completeness of several models, especially for open-source models, leading to clearer multi-region layouts and more explicit textual annotations. 
However, the improvement is not uniform: weaker models still struggle with scientific text, relation consistency, and layout fidelity. 
Among closed-source models, GPT Image 2 realizes more atoms, but also introduces more unexpected content. 
Notably, under structured prompts, Nano Banana Pro achieves a better balance among instruction faithfulness, reasoning enrichment, and semantic precision, producing substantially fewer unexpected atoms than under free-form prompts. Additional visualizations and analyses are provided in the Appendix\ref{vis}.

\vspace{-0.6cm}
\paragraph{Human Agreement Validation.} To validate the reliability of our automatic evaluation, we conduct a human-alignment study. A stratified human-alignment subset is constructed  by uniformly sampling 10 reference cases for each discipline $\times$ figure type $\times$ prompt regime cell and evaluating the corresponding outputs from all 9 T2I models, resulting in 3$\times$2$\times$2$\times$10$=$120 reference cases and 120$\times$9=1080 generated figures. Each generated figure is independently rated by three domain experts using the same atom-level rubric as our automatic judges. As shown in Tab.\ref{tab:judge_agreement}, we compare the aggregated human scores with GPT-5.4 and Qwen3.5-397B-A17B across IF, RE, and SP. We report human-human reliability using ICC, judge-human rank judge-judge agreement using Spearman's $\rho$. The high agreement across all metrics indicates that our automatic evaluation closely tracks expert judgment and remains stable across independent judge models.

\vspace{-0.5cm}
\begin{table}[htbp]
\centering
\small
\caption{
Agreement among human experts, GPT-5.4, and Qwen3.5-397B-A17B.
We report human-human reliability, human--judge rank correlation ,and
judge--judge correlation.
}
\label{tab:judge_agreement}
\resizebox{\linewidth}{!}{
\begin{tabular}{lcccc}
\toprule
\textbf{Metric} 
& \textbf{Human-Human ICC$\uparrow$} 
& \textbf{GPT-Human $\rho\uparrow$ } 
& \textbf{Qwen-Human $\rho\uparrow$ } 
& \textbf{GPT-Qwen $\rho\uparrow$ } 
\\
\midrule
IF      & 0.91 & 0.89 & 0.86 & 0.87  \\
RE      & 0.88 & 0.84 & 0.79 & 0.83\\
SP      & 0.94 & 0.88 & 0.83 & 0.85\\
\bottomrule
\end{tabular}
}
\end{table}
\vspace{-0cm}

\vspace{-0.3cm}

\section{Conclusion}
\vspace{-0.2cm}
In this work, we present \textbf{FEPBench}, a benchmark for evaluating T2I models on scientific illustration generation. 
We collect high-quality data across multiple disciplines and layout types, and provide both free-form and structured prompts. 
Based on expert-audited atom annotations, we conduct fine-grained evaluation along three dimensions: instruction faithfulness, reasoning enrichment, and semantic precision. 
Our experiments reveal the clear advantage of closed-source models, the persistent bottleneck in text generation, the general weakness in scientific reasoning, and the difficulty of balancing richness with precision. 
We further analyze how model performance is affected by layout complexity, prompt format, and atom-set complexity. 
One limitation is that the absolute number of reasoning atoms in our samples is relatively small, which may introduce variance and lead to lower reasoning-related scores. 
Developing more principled ways to evaluate scientific reasoning in visual generation remains an important direction for future work.


{
\small
\bibliographystyle{plainnat}
\bibliography{references}
}



\appendix

\section{Prompt used in MLLM}
\label{prompt}

\begin{figure}[htbp]  
\centering
\includegraphics[width=1\textwidth]{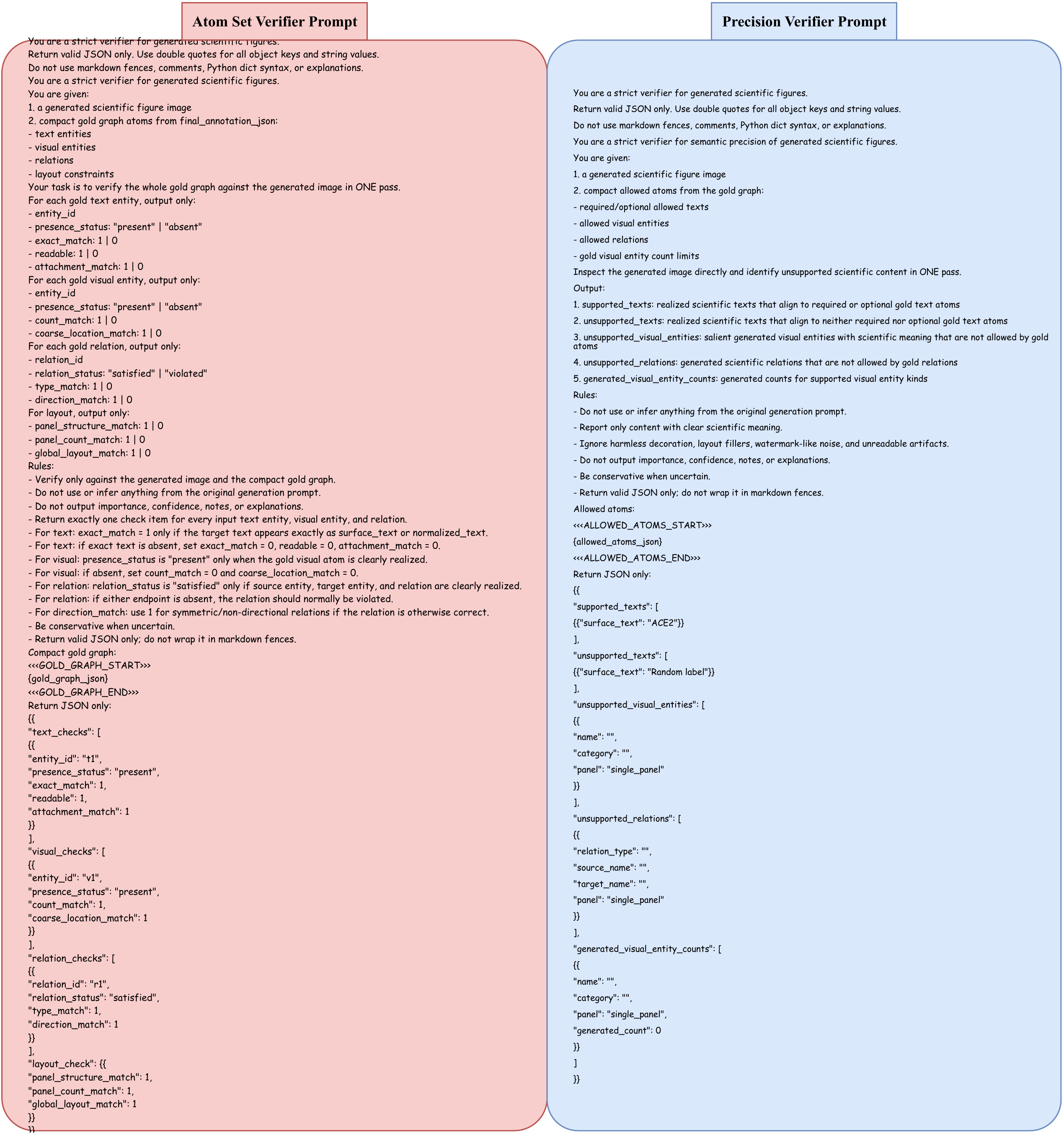} 
\caption{ Prompts for Atom Set Verifier and Precision Verifier of MLLM.} 
\label{prom}   
\end{figure}

\begin{figure}[htbp]  
\centering
\includegraphics[width=1\textwidth]{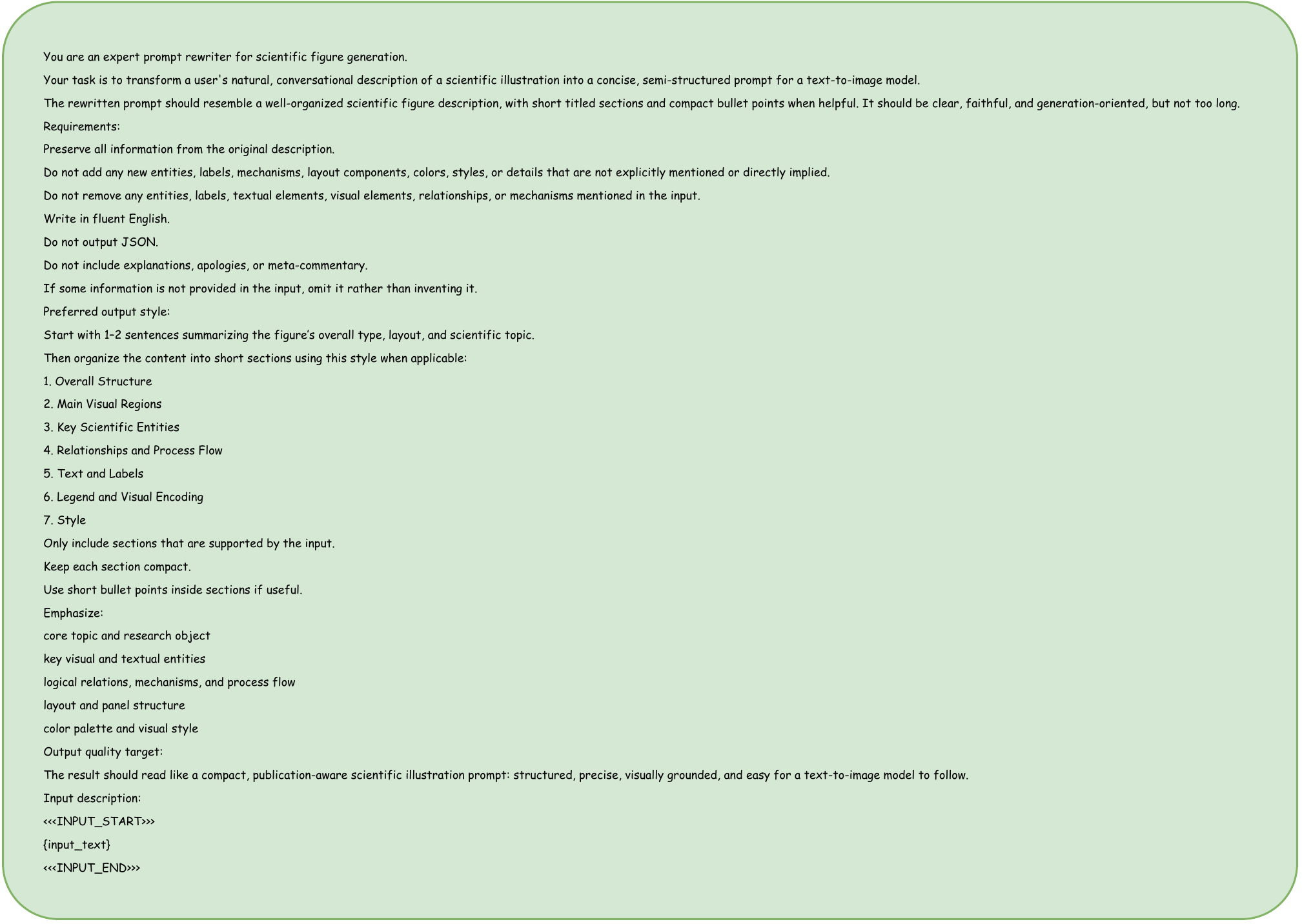} 
\caption{ Prompt used for rewriting free-form prompts into structured prompts }
\label{stru}   
\end{figure}

Here, we provide the prompts used for atom-set scoring and unexpected-atom detection during evaluation in Fig.\ref{prom}. Due to space limitations, the prompts for constructing atom sets will be included in the released code repository. The prompt used for rewriting free-form prompts into structured prompts is in Fig.\ref{stru}

\section{Statistic analysis of Results}
\label{stat}

\begin{figure}[htbp]  
\centering
\includegraphics[width=1\textwidth]{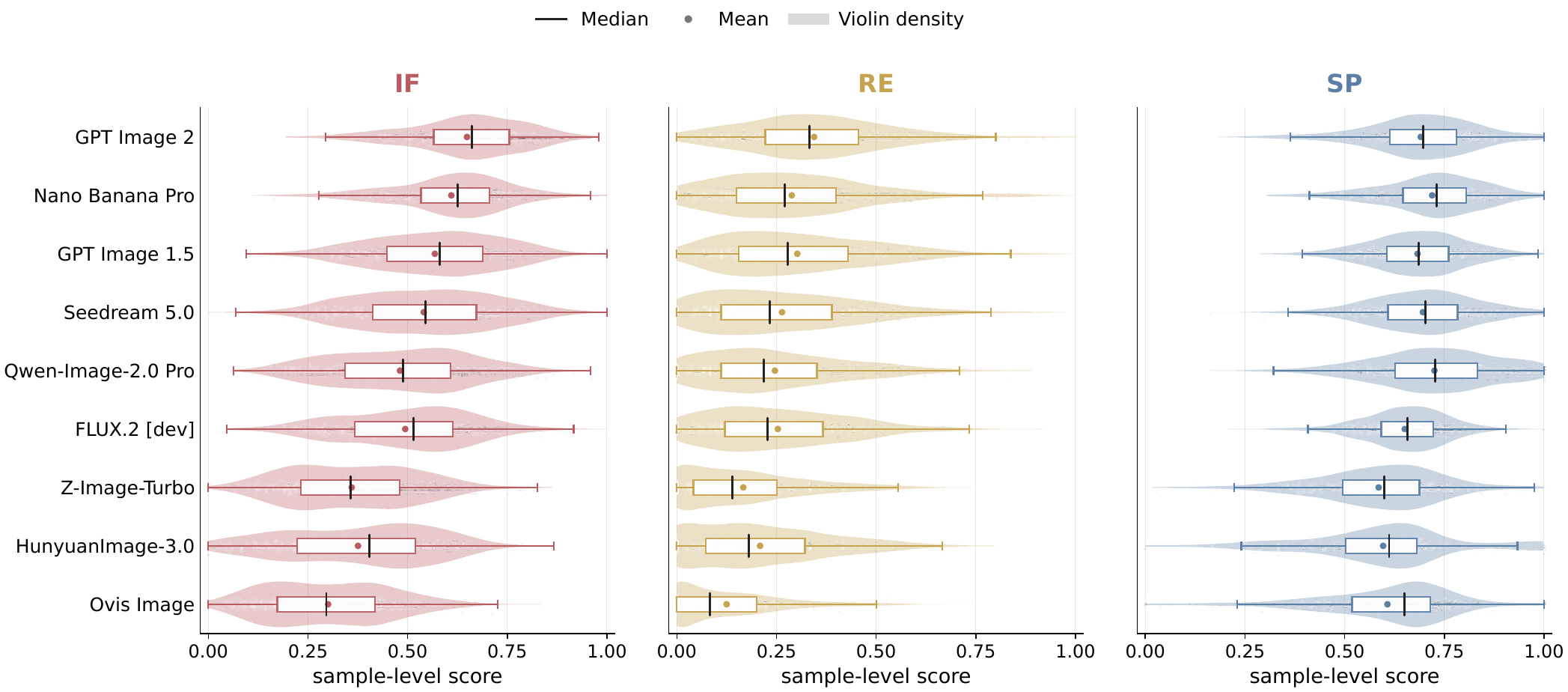} 
\caption{ Sample-level score distributions across models.
For each model, we show the distributions of Instruction Faithfulness (IF), Reasoning Enrichment (RE), and Semantic Precision (SP) over individual generated scientific illustrations.
Violin shapes represent empirical score densities, boxes indicate interquartile ranges, black ticks denote medians, and dots denote means.
} 
\label{violin}   
\end{figure}

Appendix Fig~\ref{violin} provides a sample-level view of model performance across the three primary evaluation dimensions: Instruction Faithfulness (IF), Reasoning Enrichment (RE), and Semantic Precision (SP).
Unlike aggregate tables that report only mean scores, this visualization characterizes the full distribution of scores for each model, revealing both central tendency and variability across individual scientific illustrations.
For each metric, the violin shape reflects the empirical score density, the box indicates the interquartile range, the black tick marks the median, and the dot denotes the mean.

\begin{figure}[htbp]  
\centering
\includegraphics[width=0.5\textwidth]{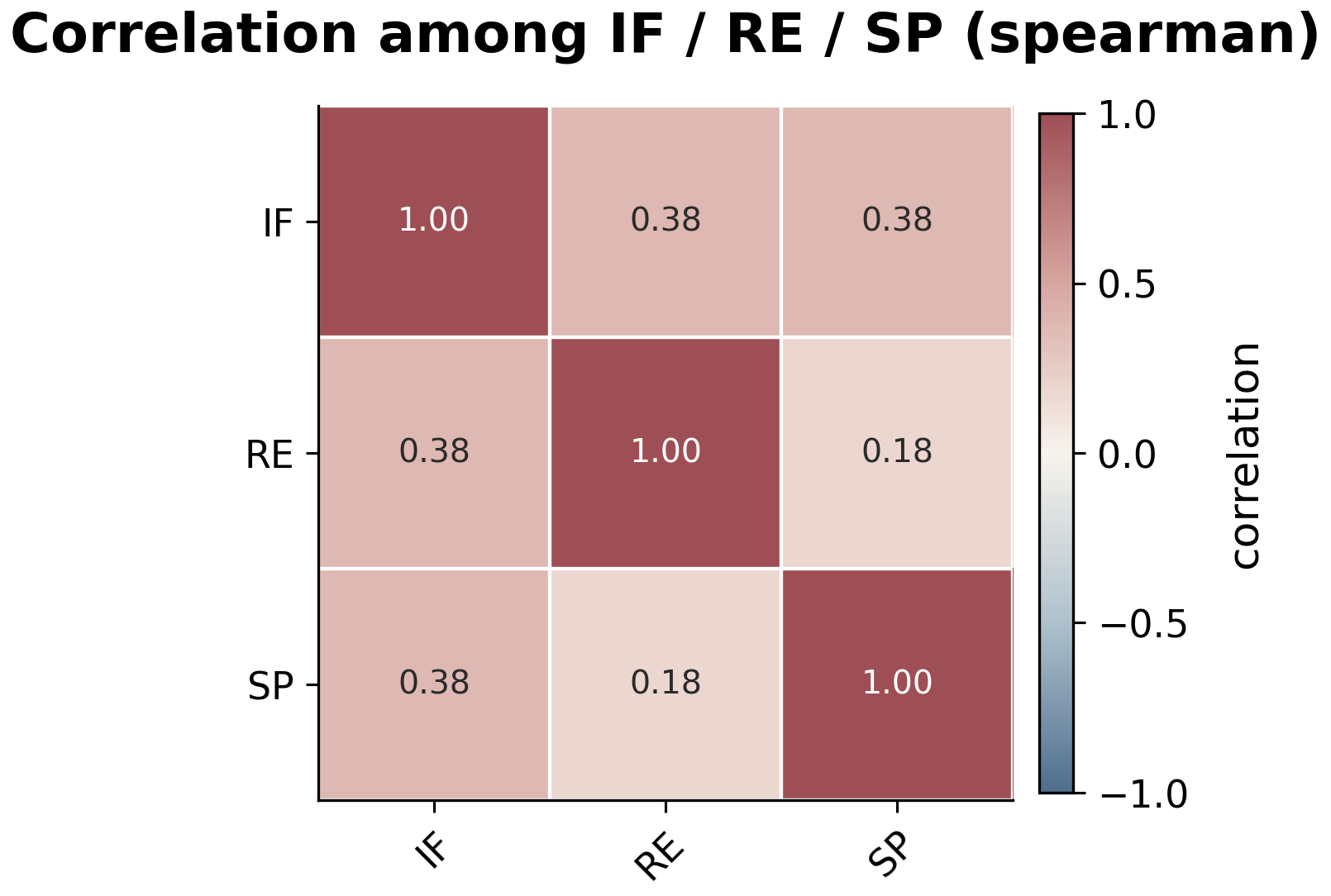} 
\caption{ Correlation among IF, RE , and SP} 
\label{corr}   
\end{figure}

\paragraph{Correlation among evaluation dimensions.}
We further analyze the pairwise correlations among the three primary metrics, as shown in Fig.~\ref{corr}.
Although some models tend to perform consistently well or poorly across IF, RE, and SP, the correlation coefficients reveal that these metrics are not strongly correlated with each other.
This indicates that the three dimensions capture complementary aspects of scientific illustration generation: instruction following, reference-level semantic enrichment, and avoidance of unsupported content.
Therefore, the proposed metric design avoids collapsing model performance into a single redundant capability and provides a more fine-grained and diagnostic evaluation.

\section{Generation Results of Model and Analysis}
\label{vis}

\begin{figure}[htbp]  
\centering
\includegraphics[width=0.8\textwidth]{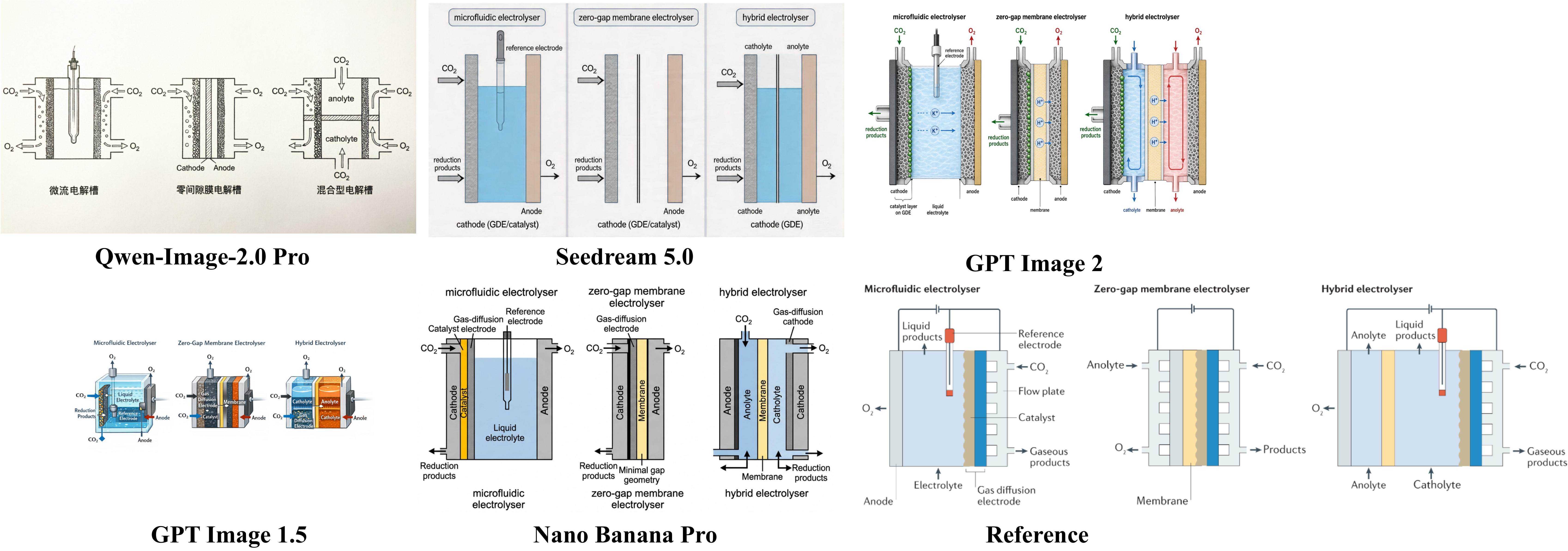} 
\caption{ Closed-source model generations on Physics and Materials tasks.} 
\label{v_phy}   
\end{figure}

\begin{figure}[htbp]  
\centering
\includegraphics[width=0.8\textwidth]{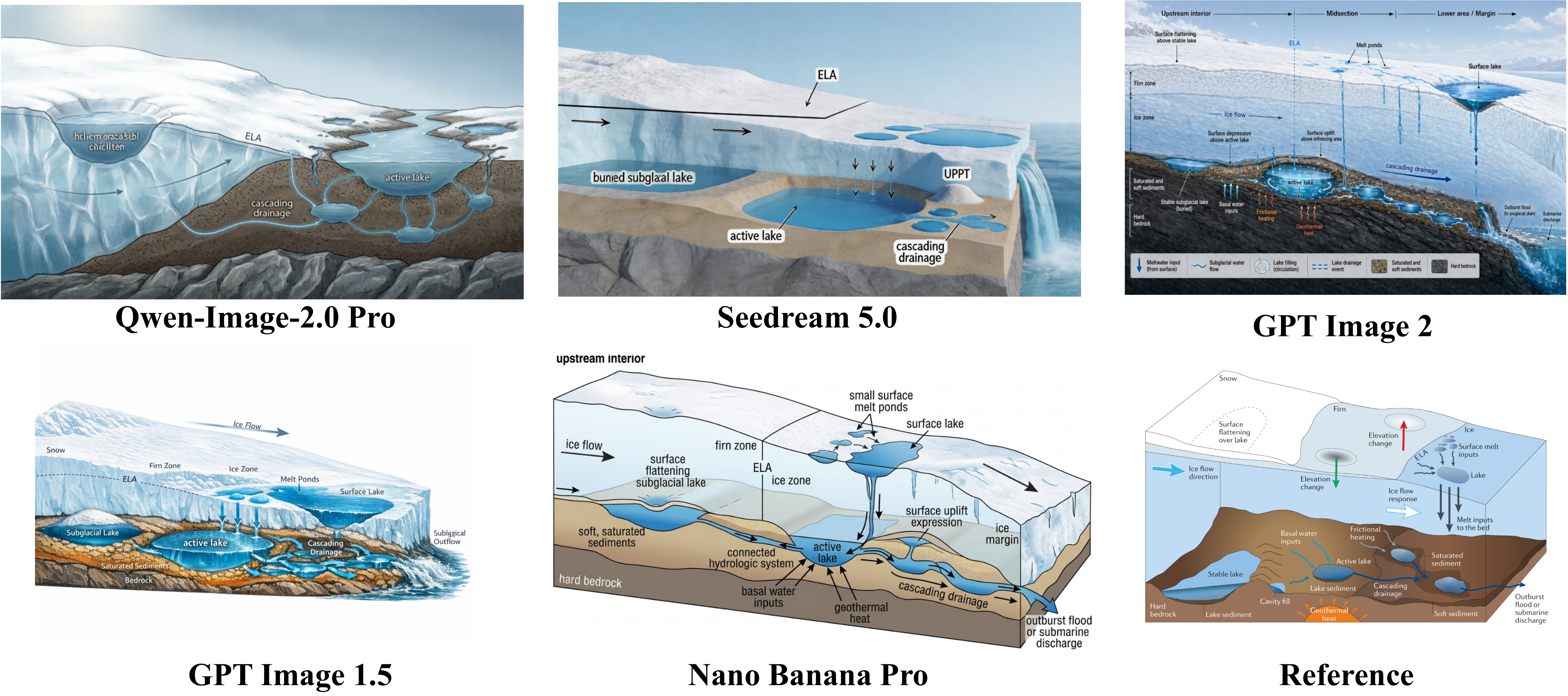} 
\caption{ Closed-source model generations on Geography and Ecology tasks.} 
\label{v_geo}   
\end{figure}

\section{Details of Models and Data}

To facilitate reproducibility, we provide comprehensive details on the dataset construction process, model configurations, generation protocols, and evaluation settings, enabling researchers to replicate our results and extend the benchmark in future studies.

\begin{table}[htbp]
\centering
\caption{
Mean numbers of instruction and reasoning atoms across benchmark subsets.
Here, $|\mathcal{A}^{ins}_{x}|$ and $|\mathcal{A}^{rea}_{x}|$ denote the number of instruction and reasoning atoms in modality $x \in \{v,t,r\}$, corresponding to visual, text, and relation atoms, respectively.
}
\label{tab:atom_statistics_by_subset}
\resizebox{\linewidth}{!}{
\begin{tabular}{llrrrrrrrrrrr}
\toprule
\multirow{2}{*}{Discipline}
& \multirow{2}{*}{Type}
& \multirow{2}{*}{$N$}
& \multicolumn{4}{c}{Ins.}
& \multicolumn{4}{c}{Rea.} \\
\cmidrule(lr){4-7} \cmidrule(lr){8-11}
& &
& $|\mathcal{A}^{ins}_{v}|$
& $|\mathcal{A}^{ins}_{t}|$
& $|\mathcal{A}^{ins}_{r}|$
& $|\mathcal{A}^{ins}|$
& $|\mathcal{A}^{rea}_{v}|$
& $|\mathcal{A}^{rea}_{t}|$
& $|\mathcal{A}^{rea}_{r}|$
& $|\mathcal{A}^{rea}|$ \\
\midrule
\multirow{2}{*}{Phys. \& Mat.}
& single & 250 & 9.40  & 11.75 & 19.57 & 40.72 & 2.15 & 4.72 & 7.69 & 14.56 \\
& multi  & 350 & 12.81 & 15.61 & 24.73 & 53.15 & 2.51 & 9.11 & 8.84 & 20.45 \\
\midrule
\multirow{2}{*}{Geo. \& Eco.}
& single & 100  & 11.74 & 16.94 & 26.28 & 54.96 & 2.49 & 5.77 & 9.49 & 17.75 \\
& multi  & 100  & 12.41 & 18.18 & 26.28 & 56.87 & 2.70 & 8.73 & 9.31 & 20.74 \\
\midrule
\multirow{2}{*}{Bio. \& Med.}
& single & 250 & 13.16 & 21.29 & 32.17 & 66.62 & 1.95 & 4.83 & 9.30 & 16.07 \\
& multi  & 250 & 14.04 & 21.93 & 32.19 & 68.16 & 2.76 & 9.65 & 9.23 & 21.63 \\
\bottomrule
\end{tabular}
}
\end{table}

\begin{table*}[t]
\centering
\small
\setlength{\tabcolsep}{5.2pt}
\renewcommand{\arraystretch}{1.08}
\caption{
Generation settings for the evaluated text-to-image models.
For API-based models, low-level sampling parameters such as steps and guidance scale are not exposed by the interface.
For open-source models, we use 50 inference steps and a guidance scale of 5.0. Test time is April 2026.
}
\label{tab:generation_settings}
\resizebox{\textwidth}{!}{
\begin{tabular}{llcccc}
\toprule
Model & Access & Generation Size & Steps & CFG / Guidance & Neg. Prompt \\
\midrule
GPT-image-2 
& API 
& 1024$\times$1024 / 1536$\times$1024 / 1024$\times$1536
& Not exposed 
& Not exposed 
& None \\

GPT-image-1.5 
& API 
& 1024$\times$1024 / 1536$\times$1024 / 1024$\times$1536
& Not exposed 
& Not exposed 
& None \\

Nano Banana Pro 
& API 
& 1920$\times$1920 / 1920$\times$1080 / 1080$\times$1920
& Not exposed 
& Not exposed 
& None \\

Seedream 5.0 
& API 
& 1920$\times$1920 / 1920$\times$1080 / 1080$\times$1920
& Not exposed 
& Not exposed 
& None \\

Qwen-Image-2.0 Pro 
& API 
& 1920$\times$1920 / 1920$\times$1080 / 1080$\times$1920
& Not exposed 
& Not exposed 
& None \\

\midrule
FLUX.2 [dev] 
& Open-source 
& 1920$\times$1920 / 1920$\times$1080 / 1080$\times$1920
& 50 
& 5.0 
& None \\

HunyuanImage-3.0 
& Open-source 
& 1024$\times$1024 / 1536$\times$1024 / 1024$\times$1536
& 50 
& 5.0 
& None \\

Z-Image-Turbo 
& Open-source 
& 1920$\times$1920 / 1920$\times$1080 / 1080$\times$1920
& 50 
& 5.0 
& None \\

Ovis-Image 
& Open-source 
& 1920$\times$1920 / 1920$\times$1080 / 1080$\times$1920
& 50 
& 5.0 
& None \\
\bottomrule
\end{tabular}
}
\end{table*}

\end{document}